# The "Room Theory": a computational model to account subjectivity into Natural Language Processing

**Carlo Lipizzi[1], Dario Borrelli[1], Fernanda de Oliveira Capela[1]**

[1] School of Systems & Enterprises, Stevens Institute of Technology, Hoboken, New Jersey, USA

Contacts: clipizzi@stevens.edu, dborrell@stevens.edu, fcapela@stevens.edu

**Abstract**

*This work introduces a novel method to consider subjectivity and general context dependency in text analysis. The proposed method takes into account subjectivity using a computational version of the Framework Theory by Marvin Minsky (1974) leveraging on text vectorization - such as Word2Vec by Mikolov et al. (2013). The embeddings created by the vectorization represent the knowledge of the context to be used to for the text analysis. Our approach is based on three components: 1. a framework/"room" representing the point of view of the individual or the collective; 2. a benchmark/set of keywords representing the criteria for the analysis; and 3. the document to be analyzed. By measuring the similarity between the vectors representing words/semantic elements, we extract the relative relevance of the elements in the benchmark for the document to be analyzed. Our method provides a way to consider the point of view of the reader of the document or the specific domain we want to use to get insights. This method could be applied to all the cases where evaluating subjectivity is relevant to understand the relative value or meaning of a text. Subjectivity is relevant to evaluate human reactions or to analyze text in a given context or domain.*

**Keywords:** Subjectivity, Text Mining, Natural Language Processing, Text Vectorization, Social Media

## 1. Introduction

Subjectivity refers to the idea that any opinion of an individual (or a collective) is shaped by its socio-cultural experience. The way one feels and reacts is affected by the social and cultural situations the subject has been exposed. Each subject has its own experience, creating their unique way to read the environment. This form of diversity affects different aspects of human life. For instance, political subjectivity is the set of thoughts, motivations, feelings that a social subject has within a society. This type of subjectivity determines the bias towards a political ideology *(Ransom, 1997)*. According to Damasio *(2018)*, subjectivity is the central constitutive element of consciousness. Without subjectivity, the individual is incapable of reflection and discernment and, therefore, is incapable of being creative.

Subjectivity is also linked to emotions, as highlighted by Harré *(1986)*, stating that emotions are the result of a social construction mechanism ("The social construction of emotions"). More recently, the anthropologist Tanya M. Luhrmann *(2006)* creates a theory of subjectivity with the help of a psychological model of emotion. Emotional responses are the consequence of the way the subject perceived an external stimulus. Due to subjectivity, this response occurs differently, depending on the subject perceiving it.

Subjectivity has an impact in a wide variety of human activities and the criteria to determine it may be applied to evaluate a more general relative relevance of a context to evaluate text. We will provide some examples of this in the conclusions.

In this paper, we use as a test case to evaluate our approach to value subjectivity in text analysis, the emotional reactions to a divisive political topic.

We use emotions as an example of a criteria for subjective classifications, being emotions a quintessential example of subjectivity. For all intents and purposes, emotions are the "what" we want to evaluate, not the "how", that is the core of this paper

In more general terms, our approach to subjectivity can be applied to all the cases when a context-dependent analysis is required. We will mention later the use of this approach to classify documents based on a given point of view.

To frame the emotions, we used the emotion classification created by Plutchik in 1980. This framework - summarized by the chart in Figure 1 – is known as the "Plutchik's wheel of emotions". According to this approach, there are eight major emotional "channels", with three levels of intensity each. Using this classification, we could define the emotions polarity according to the context where those emotions have been evaluated. Admiration/trust/acceptance may be positive in analyzing what people in a sport team are saying, negative in a radicalization analysis. Using a combination of emotions, we could also determine what Plutchik defined as "condition", such as $fear + trust = submission$.

Plutchik's wheel in Figure 1 reports 8 basic channels consisting of 3 different shades each. Between two different emotional channels, emotional conditions are also reported

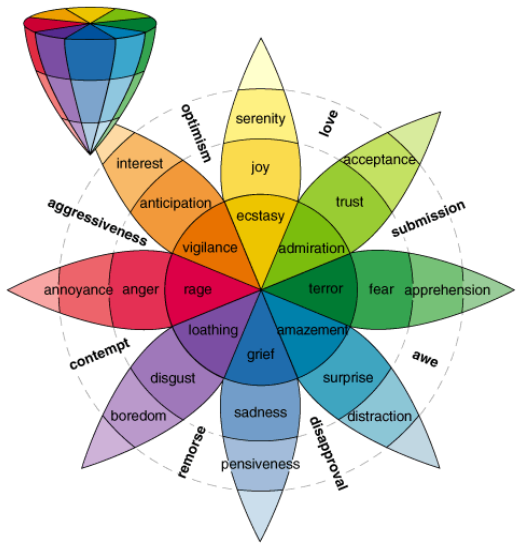

*Figure 1*

A common method to extract emotions from text is based on ontologies of language *(Shivhare & Khethawat, 2012)*. Emotions are extracted from text using the semantic similarity between words in the text and the words representing emotions, leveraging on ontology trees (e.g. Lin-similarity with WordNet). This approach is not factoring in the social and cultural differences as well as the continuous evolution of the language, leading to different perceptions of emotions.

Other works introduce the variable of subjectivity in sentiment *(Do, Prasad, Maag, & Alsadoon, 2019)*, but they do not provide a comprehensive method to detect emotions. For example, recent studies use manually annotated corpora as training set for supervised learning algorithms. Although some of these methods reach significant performances *(Mohammad & Bravo-Marquez, 2017)*, they do not take into consideration the intrinsic bias in the model from the manual annotation performed by humans with specific cultural and social backgrounds.

A recent approach to text processing is based on a vector representation of words, word embeddings *(Giatsoglou, et al., 2017)*, as input features. A particular type of vector representations of words, called Word2Vec, was introduced in 2013 by Mikolov and colleagues. Several Natural Language Processing (NLP) and Natural Language Understanding (NLU) tasks have been executed with this technique, with better results than rule-based methods. Word2Vec uses an Artificial Neural Network (ANN) to predict words belonging to a similar context by maximizing the probability that a given word appear near a set of other words. By training an ANN, they are able to use the weights of the nodes in the hidden layers to create a vector representation of the words in the original corpus. The matrix containing all the unique words in the text ("embeddings") is unique for the corpus used to train the Word2Vec model and is a numerical representation of the specific semantic domain represented by the corpus, assuming the corpus is large enough to be representative.

Using this approach, we generate domain specific embeddings ("rooms"), representing the points of view of the different entities reading the same text.

In this paper, we first provide a review of the most relevant studies on subjectivity. We then introduce our methodology and finally we present a case study to validate our approach.

For the case study, we use the evaluation of emotions as an example of the impact of subjectivity. As domain, we use a political example, based on data from the past United States presidential election (Clinton vs. Trump) to build the Word2Vec models representing the two factions of voters. Using the two models, we analyze the tweets published by Donald J. Trump on his Twitter account (@realDonaldTrump) to show how the emotional reactions of the two factions confirm our hypothesis of both subjectivity of emotional reaction and validity of our representation of this subjectivity.

## 2. Literature Review

Existing techniques on textual information processing concentrate on mining and retrieval of factual information (e.g., information retrieval, text classification, text clustering, among others). On the other hand, the processing of subjective perceptions, such as emotions and opinions, is still a developing field. Current sentiment analysis methods offer positive or negative outcome, as if the nuances of the human perception had only two polarities and not a large range of interpretations. This study addresses this issue and brings a new consideration to textual processing: different communities have different perceptions, opinions, and emotions.

### 2.1. Subjectivity

Subjectivity can be considered both as individual ("my reaction to something is different from someone else reaction") and as collective ("the experts in one area" or "a culturally homogeneous group"). For continuity with the case study, we focus our literature review primarily on the individual subjectivity.

On the individual side, there are many ways of thinking and studying subjectivity: philosophers, psychologists and theorists have approached the topic in many different ways. For the scientific community focused on natural language, the approach combining subjectivity and language by the psychoanalyst Jaques Lacan is particularly relevant. Lacan (1977) states that the unconscious is structured like a language. In particular, the author affirms that language reveals the nature of our psychology and, therefore, our vision of the world. With this concept, the author challenged the common-sense idea that language exists as a means of communication. Instead, he wants to show that language is an expression of subjectivity, in which words are not just "meaning's placeholders", but they convey subjective value depending on who is the entity that is reading, writing, listening, or speaking.

The link between subjectivity and emotions is highlighted by Harré (1986), which states that emotions, for a subject, are the result of a social construction mechanism ("The social construction of emotions"). According to the author, language and values of a society determine emotions of the individuals or groups that compose it. More recently, the anthropologist Tanya M. Luhrmann (2006) creates a theory of subjectivity, incorporating a psychological model of emotion. Watson (1919), precursor of researches behaviorism, started an evolutionary study of emotions.

The subjectivity of emotions in centered on the subject as an entity that understands, learns, classifies and evaluates. This is why a person can feel - for example - fear for an event or a situation that could be irrelevant for another: emotional responses are the consequence of the way the subject perceived an external stimulus.

### 2.2. Emotion Theories

Researchers have investigated several aspects of human emotions in order to converge to a set of commonly accepted emotion categories (Picard, 1997). *Table 1* (Aman, 2007) contains a recap of different classifications.

*Table 1*

| Tomkins | Izard | Plutchik | Ortony | Ekman |
|---------|-------|----------|--------|-------|
| Joy | Enjoyment | Joy | Joy | Happiness |
| Anguish | Sadness | Sorrow | Sadness | Sadness |
| Fear | Fear | Fear | Fear | Fear |
| Anger | Anger | Anger | Anger | Anger |
| Disgust | Disgust | Disgust | Disgust | Disgust |
| Surprise | Surprise | Surprise | Surprise | Surprise |
| Interest | Interest | Acceptance | | |
| Shame | Shame Shyness Guilt | Anticipation | | |

Plutchik developed his emotion "wheel" to illustrate the various relationships among emotions. After decades, it is still one of the main references in the field because it covers the numerous complex definitions of emotions into one image and stipulate the basic emotions as joy, sadness, fear, anger, disgust, surprise, acceptance and anticipation. It also comprehends the two dimensions for the basic emotions: valence (joy versus sadness) and arousal (anger in a lower level is annoyance, in higher level is rage). Written expression of emotion lacks gestures, tones and facial expressions, and instead relies on creative use of words for communicating emotion *(Aman, 2007)*. Some words convey emotion explicitly, while other are used to convey emotion implicitly depending on the context *(Clore, Ortony, & Foss, 1987)*.

### 2.3. Emotion Detection

As the word "affect" is commonly used in the scientific domain to refer to emotions, Piccard *(1997)* also denominates emotion recognition as Affective Computing. According to Aman *(2007)*, "recognition and classification of emotion in text can be regarded as a sub-field of sentiment analysis". Extracting from text insights on emotions may benefit many areas, like personality analysis and modeling *(Liu & Maes, 2004)*, text-to-speech synthesis *(Alm, Roth, & Sproat, 2005)*, consumer feedback analysis, Human-Computer Interaction and Affective Interfaces *(Brave & Nass, 2003)*, affective tutoring in e-learning applications *(Zhang, Zhou, Briggs, & Nunamaker Jr., 2006)*, affective communication systems *(Neviarouskaya, Prendinger, & Ishizuka, 2007)*, virtual counselling and design of agents based on emotional users preferences.

Earlier studies of emotion recognition relied on datasets that were manually annotated for emotion and were typically keyword-based, identifying the presence of an emotion based on the appearance of predetermined lexical markers.

Aman *(2007)* explores approaches for automatic detection of emotions in text using natural language processing and machine learning techniques, training classifiers using semantic resources such as WordNet Affect and Roget's Thesaurus.

Suttles and Ide **(2013)** used the eight basic emotions of Plutchik to treat the emotion classification task as a binary problem for four opposing emotion pairs. The approach applies distant supervision, which aims to overcome the need for a large set of manually labeled data to produce classifiers. There is still the need to train classifiers and opposed emotions are not considered as possibly coexisting.

## 2.4. Recent Applications

The automatization of subjective tasks is not new in Natural Language Processing. Many efficient algorithms, tools, and techniques have been developed in the past few years and can deliver reasonable results. More recent studies appear to focus on improving these existing methods or creating frameworks that combine them for a certain application.

Machine learning (ML) techniques are commonly used for subjective analysis, in particular for detection of opinion (**Jimenez-Marques, Gonzalez-Carrasco, Lopez-Cuadrado, & Ruiz-Mezcua, 2019**) and sentiment (**Pinto & Murari, 2019**). Goularte et al. (**2019**) used fuzzy rules to improve previous text summarization methods. Another study (**Li , et al., 2019**) is focused on subjective queries and databases. Wu et al. (**2019**) created an algorithm to deal with subjectivity on crowdsourced label aggregation problems.

Finally, a study from 2006 (**Lin, Wilson, Wiebe, & Hauptmann**) highlighted the need for a perspective analysis when detecting subjectivity in text. This line of study became known as stance detection and is commonly used in opinion mining, to identify if the author is in favor or against the object being analyzed (**D'Andrea, Ducange, Bechini, Renda, & Marcelloni, 2019**).

## 3 Methodology

### 3.1. Semantic Frames and Subjectivity

According to the Social Judgment Theory (Sherif & Hovland, 1961), individuals evaluate new ideas based on their social background. Social and cultural elements are part of individuals knowledge, that needs to be represented to be placed into consideration for our goal.

More specifically on knowledge representation, Marvin Minsky (1974) in his pioneer study in AI and Cognitive Science - "A Framework for Representing Knowledge" – introduced the idea of "frames". According to his work, "a frame is a data-structure for representing a stereotyped situation like being in a certain kind of living room".

We leveraged this concept to recreate "rooms" representing the semantic context for a specific social/cultural entity. Consider an instant of time t and a social entity P that interacts with the external environment using a textual content. This interaction would unfold into two processes. The first is an internalization, that corresponds to the reading/acquisition action: the entity is exposed to a textual content and the reading of this content is the action that the entity performs to start the process of internalization. The second process is the externalization: the entity produces a content and through the action of publishing this content makes it usable to other entities.

Introducing the time reference, we can collect the results of these two processes in a single corpus that contains both contents produced by the entity and the contents to which the entity has been exposed up to the instant of time t. It is necessary to define the starting time for the collection of textual contents. If the social entity is a single individual, we can refer to the date on which this individual began internalizations and externalizations (interactions). If the social entity is a collective, the starting time coincides with its first specific interaction. The social entity's socio-cultural background at time t will then be a function of the corpus which has been constructed from its first interaction up to the present temporal instant. Thus, for any collective or individual it is possible to build the corpus at a precise moment of time.

The timeline in *Figure 2* shows the interactions of a social entity P since time $t_0$. C is the corpus resulting from all textual interactions (externalization and internalization of textual content).

This allows us to compare the corpus related to different subjects and evaluate if the subjects are similar. It is also possible to evaluate how each one of these social entities perceives a new corpus in terms of emotion intensity, and this is will be detailed in the next section. At the instant t, emotions can be measured using the corpus C(t) representing the reader's point of view (that is a representation of the reader's socio-cultural background).

### 3.2. Accounting subjectivity

The proposed methodology for evaluating subjective inter-

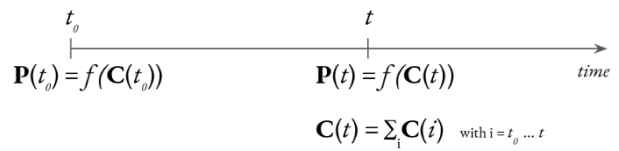

$$\mathbf{C}(t) = \Sigma_i \mathbf{C}(i) \quad \text{with } i = t_0 \dots t$$

*Figure 2*

pretation of text is summarized by the chart in *Figure 3*.

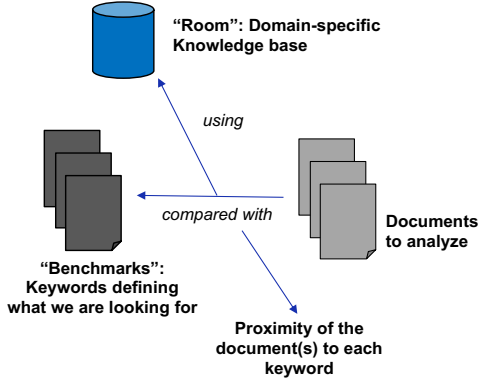

*Figure 3*

Using this methodology, we:

- Create a "room" by generating embeddings from a domain specific corpus, to represent the point of view for the analysis. This room is a computational representation of the point of view and - in a more generalized way - a computational representation of a knowledge base.

- Define a word set to be used as criteria for the analysis. This is going to be a benchmark for the comparison, like the list of emotions based on the Plutchik classification. Word set can be composed by single words or small sets of words (like in "software engineering"), referred as "chunks" or "n-grams".

- Compare words/chunks $w\_d_i$ in the incoming document (the one to be evaluated) with the words/chunks $w\_b_i$ in the benchmark, using the "room" to calculate the distance between $w\_d_i$ and $w\_b_i$

- Adding and normalize the collected similarity values for each word/chunk $w\_b_i$ in the benchmark to have an evaluation of the incoming document based on the elements in the benchmark, according to the point of view represented by the "room".

In order to provide a more accurate evaluation of the similarities, before comparing $w\_d_i$ and $w\_b_i$, we transform the words/chunks into "simsets", lists of words/chunks most similar to each $w\_d_i$ and $w\_b_i$, where the similarity is calculated by selecting the words/chunks from the "room" with the highest cosine similarity. Comparison is then performed between each element of the two lists.

### 3.3. Measuring Emotions

### 3.3.1. Transforming text into vectors

The proposed methodology is based on the distributed representation of the words - Word2Vec - introduced in the Natural Language Processing by Mikolov et al. **(2013)**, and is represented by *Figure 4*.

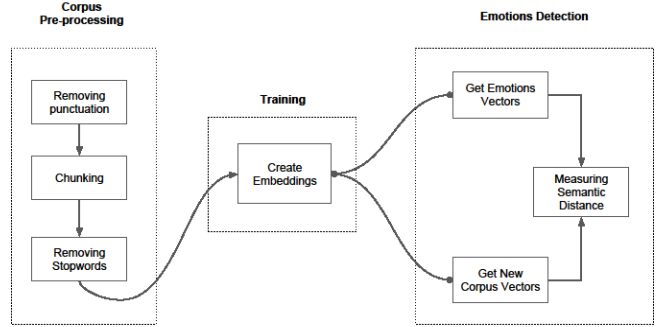

*Figure 4*

The text to be analyzed can be viewed as a string composed by n non-unique words:

$$'word_1\ word_2 \ldots word_i \ldots word_n'$$

By splitting the text string by the ' ' spaces, and cleaning the punctuation (if any), we create a single $\boldsymbol{n}$ dimensional list composed by the non-unique words contained in the text:

$$[word_1, word_2, \ldots, word_i, \ldots, word_n]$$

Using the Word2Vec method **(Mikolov, Sutskever, Chen, Corrado, & Dean, 2013)**, we assign a vector representation to each word ("embedding"), where the values defining the vectors are the output layer of the neural network used by Word2Vec and based on the probability of co-occurrence of the words in the text within a given number of words of separation. Each word is then transformed into a vector:

$$word_i \rightarrow \overrightarrow{w_i}$$

The generic $\overrightarrow{w_i}$ is an $\boldsymbol{e}$ dimensional vector, where $\boldsymbol{e}$ is equal to the embedding size used to build the Word2Vec model. The list of non-unique words can then be written as a $\boldsymbol{n}$ dimensional list of non-unique vectors each consisting of $\boldsymbol{e}$ components:

$$[\overrightarrow{w_1}, \overrightarrow{w_2}, \ldots, \overrightarrow{w_i}, \ldots, \overrightarrow{w_n}] =$$

$$= [(w_{11}, w_{12}, \ldots, w_{1e}), (w_{21}, w_{22}, \ldots, w_{2e}), \ldots,$$
$$(w_{i1}, w_{i2}, \ldots, w_{ie}), \ldots, (w_{n1}, w_{n2}, \ldots, w_{ne})]$$

If the embedding size is $\boldsymbol{e} = \boldsymbol{3}$, the words can be represented in a 3-dimensional vector space (*Figure 5*) and the list can be rewritten as follows:

$$[\overrightarrow{w_1}, \overrightarrow{w_2}, \overrightarrow{w_3}] = [(w_{11}, w_{12}, w_{13}),$$
$$(w_{21}, w_{22}, w_{23}), (w_{31}, w_{32}, w_{33})]$$

Because of the way the vectors are created by Word2Vec, the greater the probability that two words appear in the same context (meaning higher probability of co-occurrence), the higher will be the proximity of the two vector representations in the space.

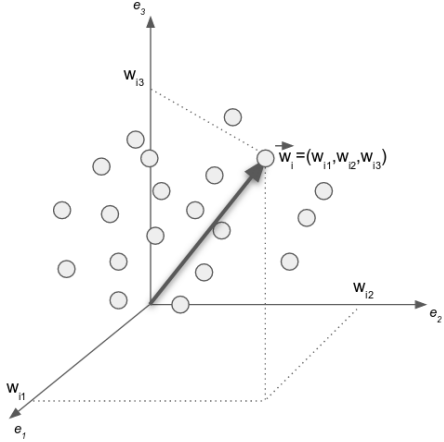

*Figure 5*

A measure that can be used to estimate the contextual proximity of two words' vectors $\overrightarrow{w_i}$ and $\overrightarrow{w_j}$, for any $e$, is the cosine similarity:

$$\overrightarrow{w_1} \overrightarrow{w_2} = \|\overrightarrow{w_1}\| \|\overrightarrow{w_2}\| \cos \theta$$

From which:

$$sim(\overrightarrow{w_1}, \overrightarrow{w_2}) = \cos \theta = \frac{\overrightarrow{w_1} \overrightarrow{w_2}}{\|\overrightarrow{w_1}\| \|\overrightarrow{w_2}\|}$$

Then, given a word, it is possible to generate a set of words contextually similar to the given word. This set, that we call "simset" in reference to the traditional synonym sets "synsets", can be defined as follows:

$$simset(\overrightarrow{w_1}) = \{\overrightarrow{w_j} : sim(\overrightarrow{w_i}, \overrightarrow{w_j}) > t\} \cup \{\overrightarrow{w_i}\}$$

where $t$ is a threshold parameter between 0 and 1.

### 3.3.2. Using vectors to evaluate emotions

We represented Plutchik's classification of emotions **(1980)** with a 8x3 matrix $E$, where each row is an emotional channel and the columns are the intensity of each emotion:

$$E = \begin{bmatrix} serenity & joy & ecstacy \\ acceptance & trust & admiration \\ apprehension & fear & terror \\ distraction & surprise & amazement \\ pensiveness & sadness & grief \\ boredom & disgust & loathing \\ annoyance & anger & rage \\ interest & anticipation & vigilance \end{bmatrix}$$

The resulting emotional conditions are included in the vector $C$ below:

$$C = \begin{bmatrix} joy + trust \\ trust + fear \\ fear + surprise \\ surprise + sadness \\ sadness + disgust \\ disgust + anger \\ anger + anticipation \\ anticipation + joy \end{bmatrix} = \begin{bmatrix} love \\ submission \\ awe \\ disapproval \\ remorse \\ contempt \\ aggressiveness \\ optimism \end{bmatrix}$$

Each emotion in the matrix $E$ can be represented as a vector using Word2Vec, generating 24 $e$ dimensional vectors, where $e$ is the embedding size. The values of the vectors are extracted as a lookup from the "room".

We can then calculate the similarity of the vector representing any word in the incoming text - also extracted as a lookup from the "room" - with each emotion. For example, given a generic '$word\ i$', whose vector is $\overrightarrow{w_i}$, we can calculate the similarity of this word with each of the 24 emotions in $E$. The result is a matrix of emotions conveyed in the given word:

$$Emotions(\overrightarrow{w_i}) =$$

$$\begin{bmatrix} sim(\overrightarrow{w_i}, serenity) & sim(\overrightarrow{w_i}, joy) & sim(\overrightarrow{w_i}, ecstacy) \\ sim(\overrightarrow{w_i}, acceptance) & sim(\overrightarrow{w_i}, trust) & sim(\overrightarrow{w_i}, admiration) \\ sim(\overrightarrow{w_i}, apprehension) & sim(\overrightarrow{w_i}, fear) & sim(\overrightarrow{w_i}, terror) \\ sim(\overrightarrow{w_i}, distraction) & sim(\overrightarrow{w_i}, surprise) & sim(\overrightarrow{w_i}, amazement) \\ sim(\overrightarrow{w_i}, pensiveness) & sim(\overrightarrow{w_i}, sadness) & sim(\overrightarrow{w_i}, grief) \\ sim(\overrightarrow{w_i}, boredom) & sim(\overrightarrow{w_i}, disgust) & sim(\overrightarrow{w_i}, loathing) \\ sim(\overrightarrow{w_i}, annoyance) & sim(\overrightarrow{w_i}, anger) & sim(\overrightarrow{w_i}, rage) \\ sim(\overrightarrow{w_i}, interest) & sim(\overrightarrow{w_i}, anticipation) & sim(\overrightarrow{w_i}, vigilance) \end{bmatrix}$$

As mentioned in 3.3.1 we can expand the granularity of the matching using "simsets" for the words. For a 300-dimensional embedding model **(Rekabsaz, Lupu, & Hanbury, 2017)**, we set a similarity threshold $t = 0.7$ (out of 1), to each word present in the simset can be associated a weight given by the similarities with $\overrightarrow{w_i}$. As an example, if we assign to the remaining two similarities random values greater than 0.7:

$$weights(\overrightarrow{w_i}) = [sim(\overrightarrow{w_i}, \overrightarrow{w_i}), sim(\overrightarrow{w_i}, \overrightarrow{w_m}), sim(\overrightarrow{w_i}, \overrightarrow{w_n})) ] =$$

$$= [1, 0.89, 0.78]$$

At this point, each emotion can be rewritten by considering the simset. For example, for joy it becomes:

$$joy(\overrightarrow{w_i}) =$$

$$= \frac{sim(\overrightarrow{w_i}, \overrightarrow{joy}) + sim(\overrightarrow{w_m}, \overrightarrow{joy}) * 0.89 + sim(\overrightarrow{w_n}, \overrightarrow{joy}) * 0.78}{|simset(\overrightarrow{w_i})|}$$

For trust, it becomes:

$$trust(\overrightarrow{w_i}) =$$

$$= \frac{sim(\overrightarrow{w_i}, \overrightarrow{trust}) + sim(\overrightarrow{w_m}, \overrightarrow{trust}) * 0.89 + sim(\overrightarrow{w_n}, \overrightarrow{trust}) * 0.78}{|simset(\overrightarrow{w_i})|}$$

Where the $|simset(\overrightarrow{w_i})|$ is the cardinality of $simset(\overrightarrow{w_i})$, which is equal to 3 in this case. Then, the resulting emotional condition will be:

$$love(\overrightarrow{w_i}) = joy(\overrightarrow{w_i}) + trust(\overrightarrow{w_i})$$

The following are some examples from sample texts. *Figure 6* is an example of emotional stacked bar chart ("emotional DNA") and the emotional conditions histogram for a given text ("terrorist attack paris"). Colors refer to those used in the Plutchik's wheel of emotions in *Figure 1*.

## 4  Case study

A case study to validate the above methodology, we analyze the emotional reactions to Trump's tweets for different groups of population.

As different groups, we selected a potential "pro Trump" and a potential "against Trump". To create the "rooms" for two different groups of population, we collected about 1.6 million tweets published between September 2016 and November 2016 by geolocated users in the United States. Tweets in this dataset are classified into two classes: those that favor the candidate Donald Trump, and those that favor the candidate Hillary Clinton.

To classify whether a tweet is belonging to the first or second class, we used the hashtag co-occurrence method proposed by Bovet et al. **(2018)**: the authors extract 4 sets of partisan hashtags used by Twitter users during the presidential race. These sets are: 1 set of hashtags in favor of Trump, 1 set of hashtags in favor of Clinton, 1 set of hashtags against Trump, 1 set of hashtags against Clinton. The 4 sets created with this approach are summarized in *Table 2*. Using this classification, we created two corpora: a "trumpers-corpus" and a "clintoners-corpus".

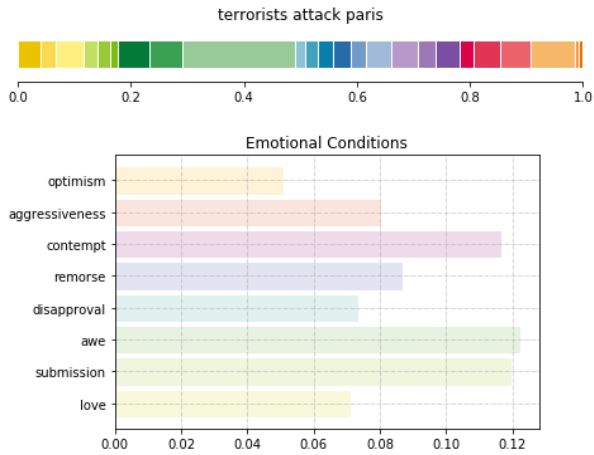

*Figure 6*

*Table 2*

| Entity | Set | Hashtags |
|---|---|---|
| Trump- ers | Pro- Trump | #trump2016, #trump16, #makeamericagreatagain, #maga, #trumppence16,#trumptrain, #presidenttrump, #makeamericasafeagain, #democratsfortrump, #vetsfortrump, #women4trump, #gays4trump, #democrats4trump, #latinos4trump, #blacks4trump, #buildthewall, #votetrump2016, #alwaystrump, #bikers- fortrump, #makeamericaworkagain, #trumpiswithyou, #onlytrump, #heswithus, #trumpcares, #votegop |
| Trump- ers | Anti- Clinton | #neverhillary, #imnotwithher, #crookedhillary, #nevereverhillary, #nomoreclintons, #stophillary, #kilary, #clintoncrimefoundation, #hillno, #dropouthillary, #riskyhillary, #clin- toncorruption, #notwither, #hillary4jail, #deletehillary, #hillarylies, #hypocritehillary, #iwillneverstandwithher, #crookedclinton, #crookedclintons, #lyinghillary, #hillaryliesmatter, #hillariedpeopledied |
| Clin- toners | Pro-Clin- ton | #hillary2016, #imwither, #strongertogether, #vote4hillary, #imwithhillary, #clinton- kaine2016, #hillarysopresidential, #hillarystrong, #uniteblue, #voteblue, #sheswithus, #votehillary, #madampresident, #yeswekaine, #welovehillary, #itrusther, #istrusthillary, |

| | | #estoyconella, #repubblicans4hillary, #bluewave2016, #hillstorm2016, #hillaryforpr, #hillaryforamerica, #hillarysoqualified, hillaryforpresident |
| --- | --- | --- |
| Clintoners | Anti-Trump | #nevertrump, #dumpthetrump, #crybabytrum, #trumpthefraud, #lyingtrump, #stoptrump, #dirtydonald, #crookeddonald, #lyintrump, #nevertrumppence, #boycotttrump, #lyindonald, #lovetrumpshates, #notrumpanytime, #defeattrump, #weakdonald, #sleazydonald, #chickentrump. #loserdonald, #losertrump, #showusyourtaxes, #antitrump, #freethedelefates, #stoptrump, #traitfortrump |

We cleaned and pre-processed the corpora using the pipeline represented by *Figure 4*. The two corpora have been used to train two Word2Vec models using a Python library – gensim – with the skip-gram algorithm **(Mikolov, Sutskever, Chen, Corrado, & Dean, 2013)**, an embedding size equal to 300, window size equal to 5, and minimum number of word counts of 2

To analyze how the two groups perceive emotions, we used a dataset containing all tweets published by @realDonaldTrump, official account of Donald J. Trump. For each tweet, the emotions perceived by the two groups were calculated using the two different Word2Vec models trained with the partisan textual contents ("trumpers-corpus" and "clintoners-corpus").

Tweets by @realDonaldTrump Twitter account have been cleaned and pre-processed, same as for tweets from the groups of supporters.

Figure 7 shows the comparison between emotions for the two groups, where red is for trumpers and blue for clintoners. The lighter is the cloud the higher is the number of datapoints concentrated in that area. The difference in mean values of emotions expresses the emotional subjectivity

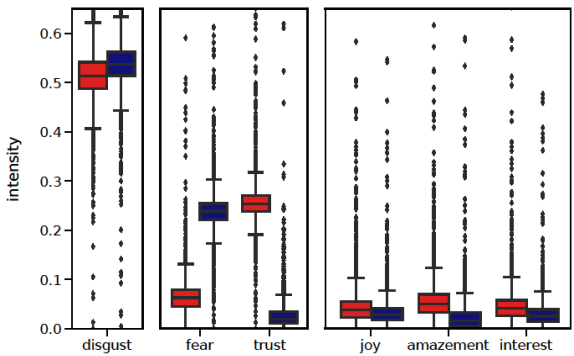

*Figure 7*

To better evaluate the results, they have been clustered, using a measure of emotional polarization **(Primario, Borrelli, Iandoli, Zollo, & Lipizzi, 2017; Morales, Borondo, Losada, & Benito, 2015)** for each emotional channel. We calculated the polarization $P$ that takes into account the distance between the two peaks of distributions (of both supporters' group) and their population in terms of number of tweets published by each group of supporters:

$$P = \left(1 - \frac{|Population(T) - Population(C)|}{Population(T) + Population(C)}\right) * d$$

Where $d = |avg(E_T) - avg(E_C)|$ is the absolute value of the difference between the average for the emotion $E$ perceived by the $T$ (trumpers) and the average for the same emotion $E$ perceived by the population $C$ (clintoners). $Population(X)$ is the total number of tweets of a group $X$.

Emotions inducing a polarizing behavior, are the first three in *Table 3*. The table aggregates results and includes values of $d$ and $P$.

*Table 3*

| Emotion | Trumpers | Clintoners | $d$ | $P$ |
| --- | --- | --- | --- | --- |
| | $avg(E_T)$ | $avg(E_C)$ | | |
| Trust | 0.256522 | 0.028836 | 0.222219 | 0.217215 |
| Fear | 0.068418 | 0.238126 | 0.165067 | 0.161353 |
| Anger | 0.041604 | 0.155071 | 0.114431 | 0.111854 |
| Amazement | 0.059374 | 0.030679 | 0.033577 | 0.032821 |
| Disgust | 0.516265 | 0.536175 | 0.010235 | 0.010004 |
| Interest | 0.049609 | 0.033509 | 0.017101 | 0.016716 |
| Joy | 0.048026 | 0.036052 | 0.004957 | 0.004845 |
| Sadness | 0.046214 | 0.037204 | 0.010704 | 0.010498 |

Using this metric, we considered the following two clusters of emotions:

$$A = \{Anger, Fear, Trust\}$$

$$B = \{Disgust, Amazement, Interest, Joy, Sadness\}$$

*Figure 8* and *Figure 9* show the distribution of values for the different clusters and emotions. On the left graphs, the lighter is the cloud the higher is the number of datapoints concentrated in that area. On the right side, the same emotions have been visualized using the pair plots, which show red distributions for "trumpers" and blue distributions for "clintoners".

*Figure 8* contains the results for cluster *A*. In fact, the intensity of Trust, which is the most polarizing emotion (0.21) is biased towards trumpers group as we are analyzing Trump's tweets. As contrary, Fear and Anger are biased towards clintoners group.

*Figure 9* contains the results for cluster *B*, composed by emotions that seems to be less polarizing than the emotions in cluster A. Those are more generic categories of emotions like Joy, Sadness but even some unexpected emotions as Disgust. In particular, Disgust has the highest intensity for both the communities of supporters with a slight advantage for clintoners (+0.02). Amazement and Interest have low values of emotional polarization (0.03 and 0.01) but both are biased towards trumpers in terms of intensity, which still confirm our hypothesis.

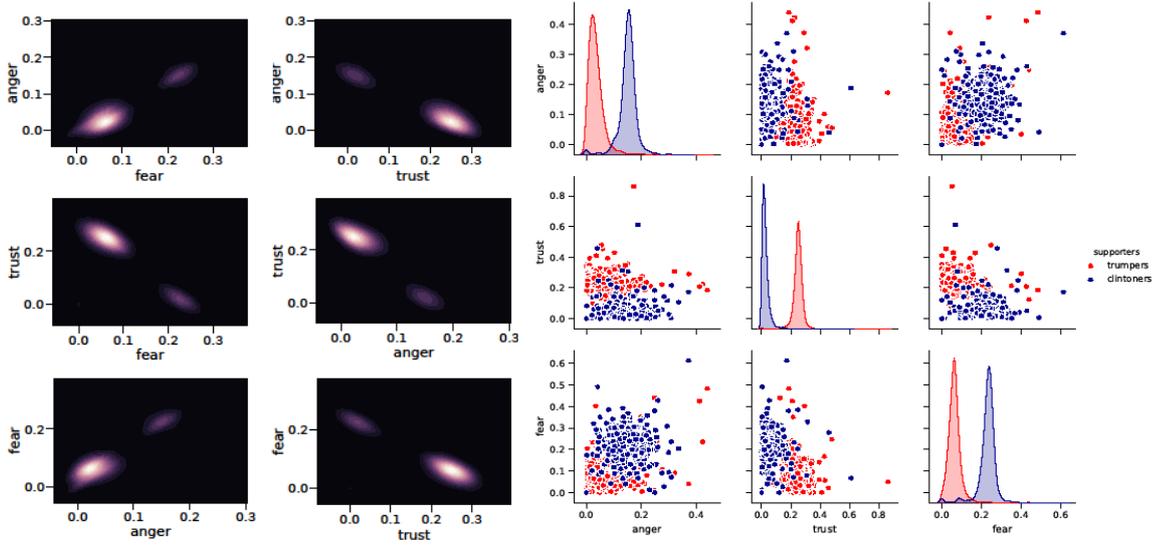

*Figure 8*

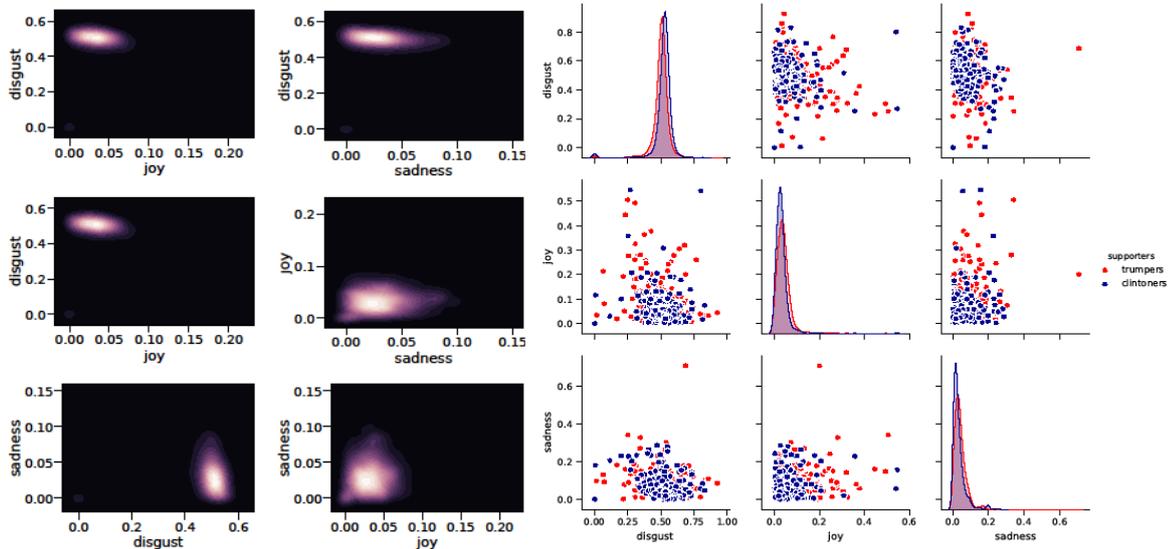

*Figure 9*

## 5 Discussion

We selected and classified two different categories of users' text to create two different perspectives from which our methodology measures the emotions. We found that by training two different Word2Vec models, with two differently biased political corpora, we obtain a distinction in intensity of emotions that confirms the hypothesis of emotional subjectivity that we made. In our experiment, the distinction is particularly significant for certain types of emotions such as Anger, Fear, and Trust. In addition, the choice of analyzing text of @realDonaldTrump tweets helped us validate the correctness of our method. In fact, emotions like Trust are perceived with greater intensity by the trumpers group, while emotions as Anger have a higher intensity for the clintoners. Anger, Trust, Fear show signals of polarization.

## 6. Conclusion

In this paper, we presented a novel approach to detect subjectivity in text corpora and applied it to one of the most subjective areas of human life, that is emotion perception.

Unlike other studies that provide a general method for measuring emotions in a text, we intend to contribute with an approach that uses the point of view of the social entity (individual or collective) that reads the text. Our approach is essentially different from what is already present, both from a methodological and interpretative perspective. Leveraging theories from Psychology **(Plutchik, 1980)**, Cognitive Sciences **(Minsky, 1974)**, Social Sciences **(Sherif & Hovland, 1961)**, and recent text mining approaches **(Mikolov, Sutskever, Chen, Corrado, & Dean, 2013)** we have built a framework able to:

o   Create a point of view ("room") to analyze a text.

o   Measure the basic emotions perceived using that point of view.

The proposed method has generated interesting results on the case study. Using two groups of individuals who interact online with different political ideas, we have found that our method produces emotional differences in the two groups, and these emotional differences are also oriented towards the political bias of the analyzed text, thus validating our method.

The application to emotion detection can be relevant in marketing, finance, politics, psychology and social science studies, providing elements to understand people's reactions to text beyond traditional sentiment analysis.

We used the "room theory" as represented by *Figure 3* with different benchmarks to extract numerical values from incoming documents.

In particular, we used it to create a Decision Support System to help decision maker to determine the most appropriate investments in technologies to better compete with their competitors. The source of information was streams of news, patents and papers related to technology and the risk was calculated using the decision maker "room". We then created a risk panel that we visualized in an interactive mode.

We also used the same approach to determine emerging and coming technologies, using as benchmark a list of current technologies, as incoming documents papers, news, patents, blogs and calculating distances using a "room" created from text related to technologies.

Another application was in determining the most appropriate contract type for given purchasing requests. In this case, the "room" was created from a large corpus representing the contracting officer knowledge base and the benchmark the list of characteristics of the different contract types.

This method could be applied to all the cases where evaluating subjectivity is relevant to understand the relative value or meaning of a text, such as emotion, sentiment, and opinion mining, language translation, text summarization, topic labeling, amongst others. Subjectivity is not limited to human reactions, but it could be used to provide a text with an interpretation related to a given domain.